\documentclass[11pt,letterpaper]{article}
\pdfoutput=1
\usepackage{acl2016}
\usepackage{times}
\usepackage{url}
\usepackage{latexsym}
\usepackage{multirow}

\usepackage{mdwlist}
\usepackage{amsmath}
\usepackage{amssymb}
\usepackage{amsfonts}
\usepackage{booktabs}
\usepackage{textcomp}
\usepackage{color}
\usepackage{graphicx}
\usepackage{fancyvrb}
\usepackage{multirow}
\usepackage{tabularx}
\usepackage{umoline}
\usepackage{caption}
\usepackage{subcaption}
\usepackage{footnote}
\usepackage{minipage}
\usepackage[font={small}]{caption}
\usepackage{soul}

\aclfinalcopy 
\def\aclpaperid{577} 

\definecolor{green}{rgb}{0.1,0.5,0.1}
\definecolor{red}{rgb}{1.0,0.2,0.2}
\definecolor{blue}{rgb}{0.0,0.0,1.0}

\newcommand{\be}{\begin{equation}}
\newcommand{\ee}{\end{equation}}

\title{Stack-propagation: Improved Representation Learning for Syntax}

\author{Yuan Zhang\thanks{~~Research conducted at Google.}\\
	    CSAIL, MIT\\
	    Cambridge, MA 02139, USA\\
	    {\tt yuanzh@csail.mit.edu}
	  \And
	David Weiss\\
  	Google Inc\\
  	New York, NY 10027, USA\\
  {\tt djweiss@google.com}}

\date{}

\begin{document}
\maketitle

\begin{abstract}
  Traditional syntax models typically leverage part-of-speech (POS) information by
constructing features from hand-tuned templates. 
We demonstrate that a better approach is to utilize POS tags as a {\em
  regularizer} of learned representations. 
We propose a simple method for learning a stacked pipeline of models which we
call ``stack-propagation.''
We apply this to dependency parsing and tagging, where we use the hidden layer
of the tagger network as a representation of the input tokens for the parser. 
At test time, our parser does not require predicted POS tags.
On 19 languages from the Universal Dependencies, our method is 1.3\% (absolute)
more accurate than a state-of-the-art graph-based approach and 2.7\% more
accurate than the most comparable greedy model. 


\end{abstract}

\newcommand{\bX}[0]{\mathbf{X}}
\newcommand{\bx}[0]{\mathbf{x}}
\newcommand{\bh}[0]{\mathbf{h}}


\section{Introduction}

In recent years, transition-based dependency parsers powered by neural network
scoring functions have dramatically increased the state-of-the-art in terms of
both speed and accuracy
\cite{chen2014fast,albert-etAl:2015:EMNLP,weiss-etAl:2015:ACL}.
Similar approaches also achieve state-of-the-art in other NLP tasks, such as
constituency parsing \cite{DurrettKlein2015} or semantic role labeling
\cite{fitzgerald15:srl}.
These approaches all share a common principle: replace hand-tuned conjunctions
of traditional NLP feature templates with continuous approximations learned by
the hidden layer of a feed-forward network.

However, state-of-the-art dependency parsers depend crucially on the use of
predicted part-of-speech (POS) tags. 
In the pipeline or {\em stacking} \cite{wolpert1992stacked} method, these are
predicted from an independently trained tagger and used as features in the
parser. 
However, there are two main disadvantages of a pipeline: (1) errors from the POS
tagger cascade into parsing errors, and (2) POS taggers often make mistakes
precisely because they cannot take into account the syntactic context of a parse
tree. 
The POS tags may also contain only coarse information, such as when using the
universal tagset of \newcite{petrov2011universal}.


One approach to solve these issues has been to avoid using POS tags during
parsing, e.g. either using semi-supervised clustering instead of POS tags
\cite{koo2008simple} or building recurrent representations of words using neural
networks \cite{dyer2015transition,BallesterosDS15}.
However, the best accuracy for these approaches is still achieved by running a
POS tagger over the data first and combining the predicted POS tags with
additional representations.
As an alternative, a wide range of prior work has investigated jointly modeling
both POS and parse trees
\cite{li2011joint,hatori2011incremental,bohnet2012transition,qian2012joint,wang2014joint,li2014joint,zhang2015randomized,albert-etAl:2015:EMNLP}.
However, these approaches typically require sacrificing either efficiency or
accuracy compared to the best pipeline model, and often they simply
re-rank the predictions of a pipelined POS tagger.

\begin{figure}
  \centering
  \includegraphics[width=1.0\linewidth]{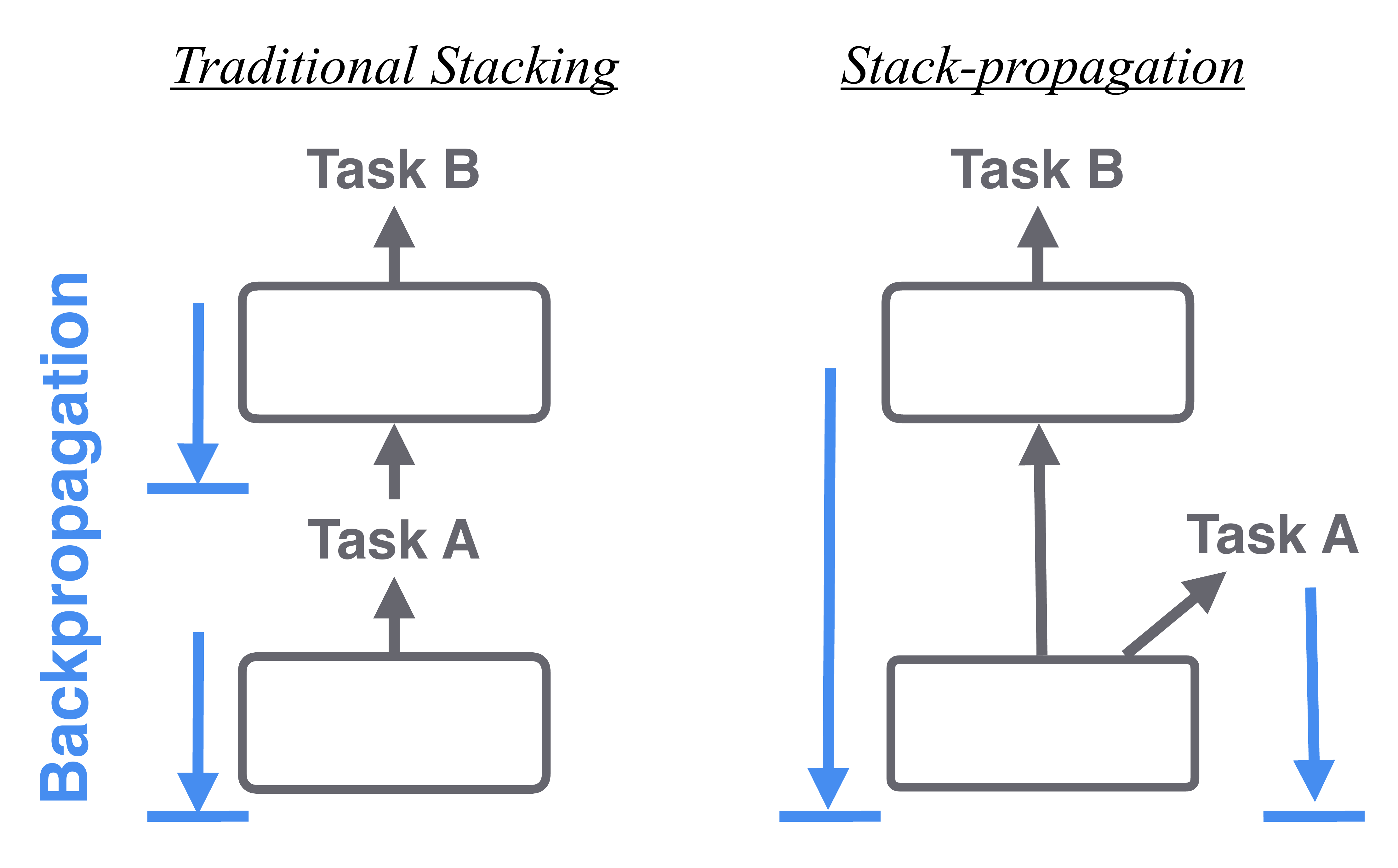}
  \caption{
    Traditional stacking (left) vs. Stack-propagation (right). 
    Stacking uses the output of Task A as features in Task B, and does not allow
    backpropagation between tasks. 
    Stack-propagation uses a continuous and differentiable link between Task A
    and Task B, allowing for backpropagation from Task B into Task A's
    model. Updates to Task A act as {\em regularization} on the model for Task
    B, ensuring the shared component is useful for both tasks.}
  \label{fig:multitask}
\end{figure}

In this work, we show how to improve accuracy for both POS tagging and parsing
by incorporating stacking into the architecture of a feed-forward network.
We propose a continuous form of stacking that allows for easy backpropagation
down the pipeline across multiple tasks, a process we call ``stack-propagation''
(Figure \ref{fig:multitask}).
At the core of this idea is that we use POS tags as {\em regularization} instead
of {\em features}.

\begin{figure*}[t]
\centering
\includegraphics[width=\linewidth]{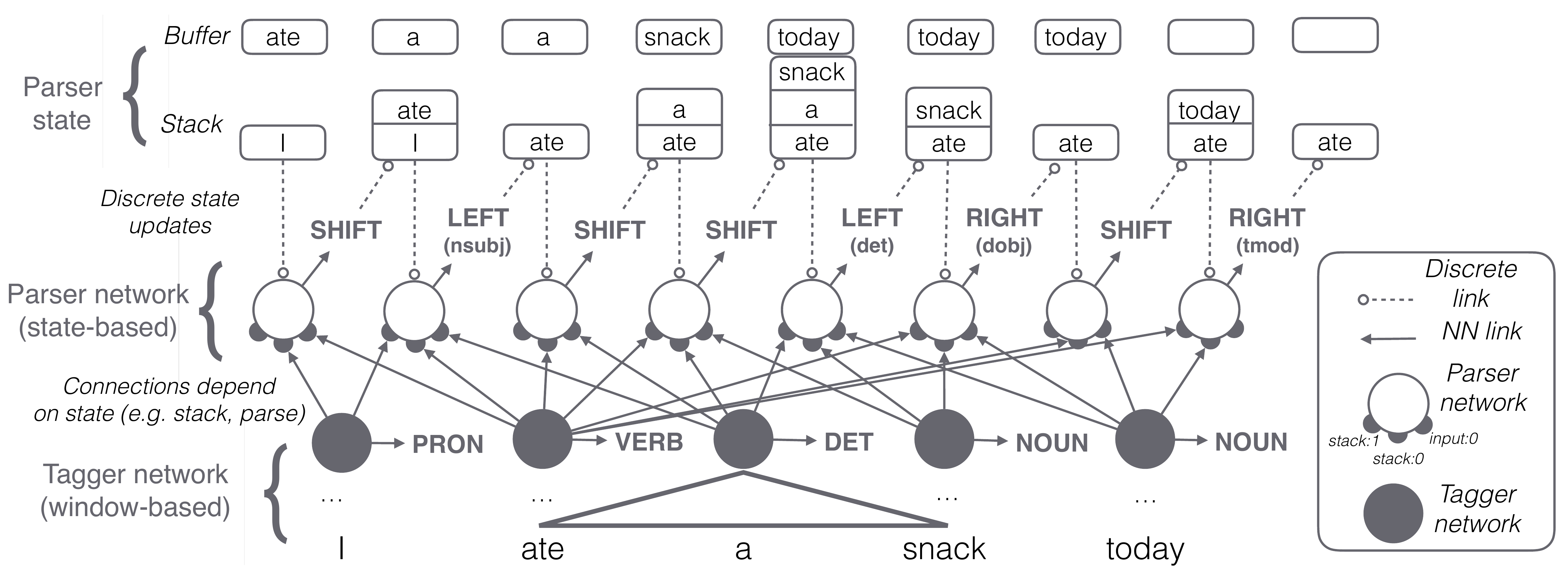}
\caption{
    Detailed example of the stacked parsing model. {\em
    Top:} The discrete parser state, consisting of the stack and the buffer, is
  updated by the output of the parser network. In turn, the feature templates
  used by the parser are a function of the state. In this example, the parser
  has three templates, {\tt stack:0}, {\tt stack:1}, and {\tt input:0}. {\em
    Bottom:} The feature templates create many-to-many connections from the
  hidden layer of the tagger to the input layer of the parser. For example, the
  predicted root of the sentence (``ate'') is connected to the 
  input of most parse decisions. At test time, the above structure is
  constructed dynamically as a function of the parser output. Note also that the
  predicted POS tags are not directly used by the parser.}
\label{fig:unrolled}
\vspace{-0.5em}
\end{figure*}

\begin{figure*}[t]
    \centering
    \begin{minipage}{0.32\textwidth}
        \centering
        \includegraphics[width=\textwidth]{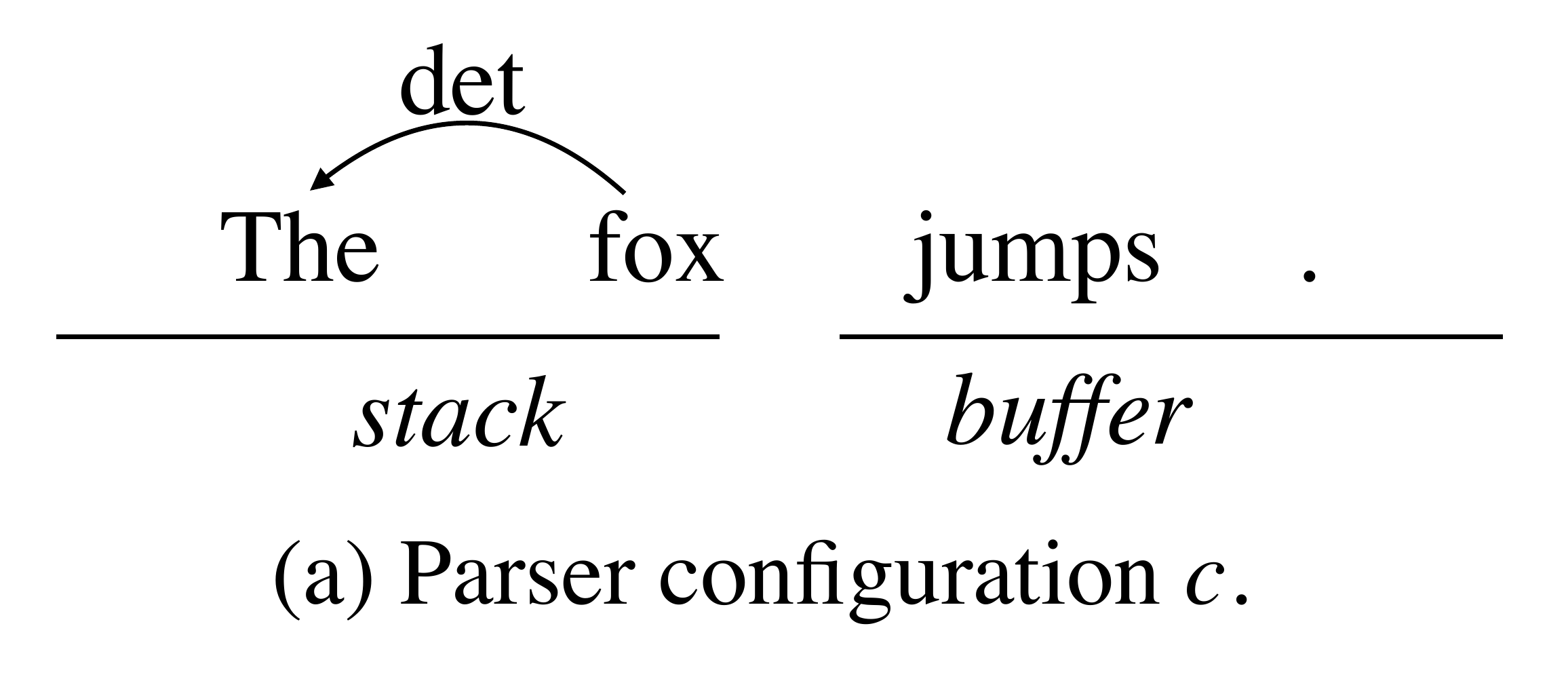}
        \caption{\small{A schematic for the \textsc{Parser} stack-propagation
            update. {\em a}: Example parser configuration $c$ with corresponding
            stack and buffer.
            {\em b}: Forward and backward stages for the given single
            example. During the forward phase, the tagger networks compute
            hidden activations for each feature template
            (e.g. \textit{stack}$_0$ and \textit{buffer}$_0$), and activations
            are fed as features into the parser network. For the backward
            update, we backpropagate training signals from the parser network
            into each linked tagging example.}}
        \label{fig:schema}
    \end{minipage}~~%
    \begin{minipage}{0.68\textwidth}
        \centering
        \includegraphics[width=\textwidth]{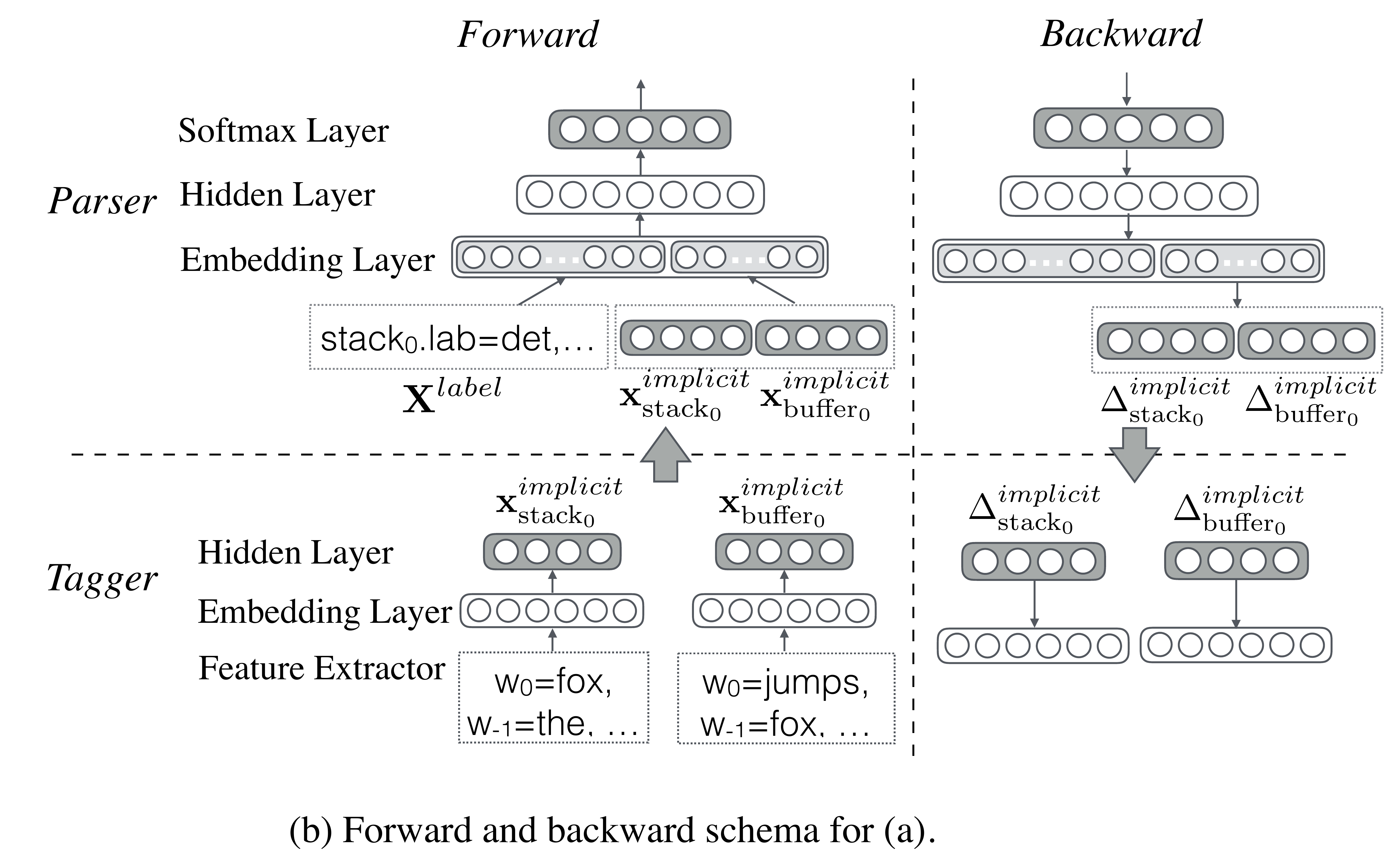}
    \end{minipage}
\vspace{-0.5em}
\end{figure*}

Our model design for parsing is very simple: we use the hidden layer of a
window-based POS tagging network as the representation of tokens in a greedy,
transition-based neural network parser.
Both networks are implemented with a refined version of the feed-forward network
(Figure \ref{fig:basic_unit}) from \newcite{chen2014fast}, as described in
\newcite{weiss-etAl:2015:ACL}.
We link the tagger network to the parser by translating traditional feature
templates for parsing into feed-forward connections from the tagger to the
parser (Figure \ref{fig:unrolled}).
At training time, we unroll the parser decisions and apply stack-propagation by
alternating between stochastic updates to the parsing or tagging objectives
(Figure \ref{fig:schema}).
The parser's representations of tokens are thus regularized to be individually
predictive of POS tags, even as they are trained to be useful for parsing when
concatenated and fed into the parser network.
This model is similar to the multi-task network structure of
\newcite{collobertJMLR2011}, where \newcite{collobertJMLR2011} shares a hidden
layer between multiple tagging tasks. The primary difference here is that we
show how to unroll parser transitions to apply the same principle to tasks with
fundamentally different structure.

The key advantage of our approach is that at test time, we do not require
predicted POS tags for parsing.
Instead, we run the tagger network up to the hidden layer over the entire
sentence, and then dynamically connect the parser network to the tagger network
based upon the discrete parser configurations as parsing unfolds.
In this way, we avoid cascading POS tagging errors to the parser.
As we show in Section \ref{sec:discussion}, our approach can be used in
conjunction with joint transition systems in the parser to improve both POS
tagging as well as parsing.
In addition, because the parser re-uses the representation from the tagger, we
can drop all lexicalized features from the parser network, leading to a compact,
faster model. 

The rest of the paper is organized as follows.
In Section \ref{sec:stacked}, we describe the layout of our combined
architecture.
In Section \ref{sec:learning}, we introduce stack-propagation and show how we
train our model.
We evaluate our approach on 19 languages from the Universal Dependencies
treebank in Section \ref{sec:experiments}.
We observe a $>$2\% absolute gain in labeled accuracy compared to
state-of-the-art, LSTM-based greedy parsers \cite{BallesterosDS15} and a $>$1\%
gain compared to a state-of-the-art, graph-based method \cite{lei2014low}.
We also evaluate our method on the Wall Street Journal, where we find that our
architecture outperforms other greedy models, especially when only coarse POS
tags from the universal tagset are provided during training.
In Section \ref{sec:discussion}, we systematically evaluate the different
components of our approach to demonstrate the effectiveness of stack-propagation
compared to traditional types of joint modeling.
We also show that our approach leads to large reductions in cascaded errors from
the POS tagger.

We hope that this work will motivate further research in combining traditional
pipelined structured prediction models with deep neural architectures that learn
intermediate representations in a task-driven manner.
One important finding of this work is that, even {\em without} POS tags, our
architecture outperforms recurrent approaches that build custom word
representations using character-based LSTMs \cite{BallesterosDS15}.
These results suggest that learning rich embeddings of words may not be as
important as building an intermediate representation that takes multiple
features of the {\em surrounding} context into account.
Our results also suggest that deep models for dependency parsing may not
discover POS classes when trained solely for parsing, even when it is fully
within the capacity of the model.
Designing architectures to apply stack-propagation in other coupled NLP tasks
might yield significant accuracy improvements for deep learning.

\section{Continuous Stacking Model}
\label{sec:stacked}

\begin{figure}[t]
  \centering
  \includegraphics[width=0.5\textwidth]{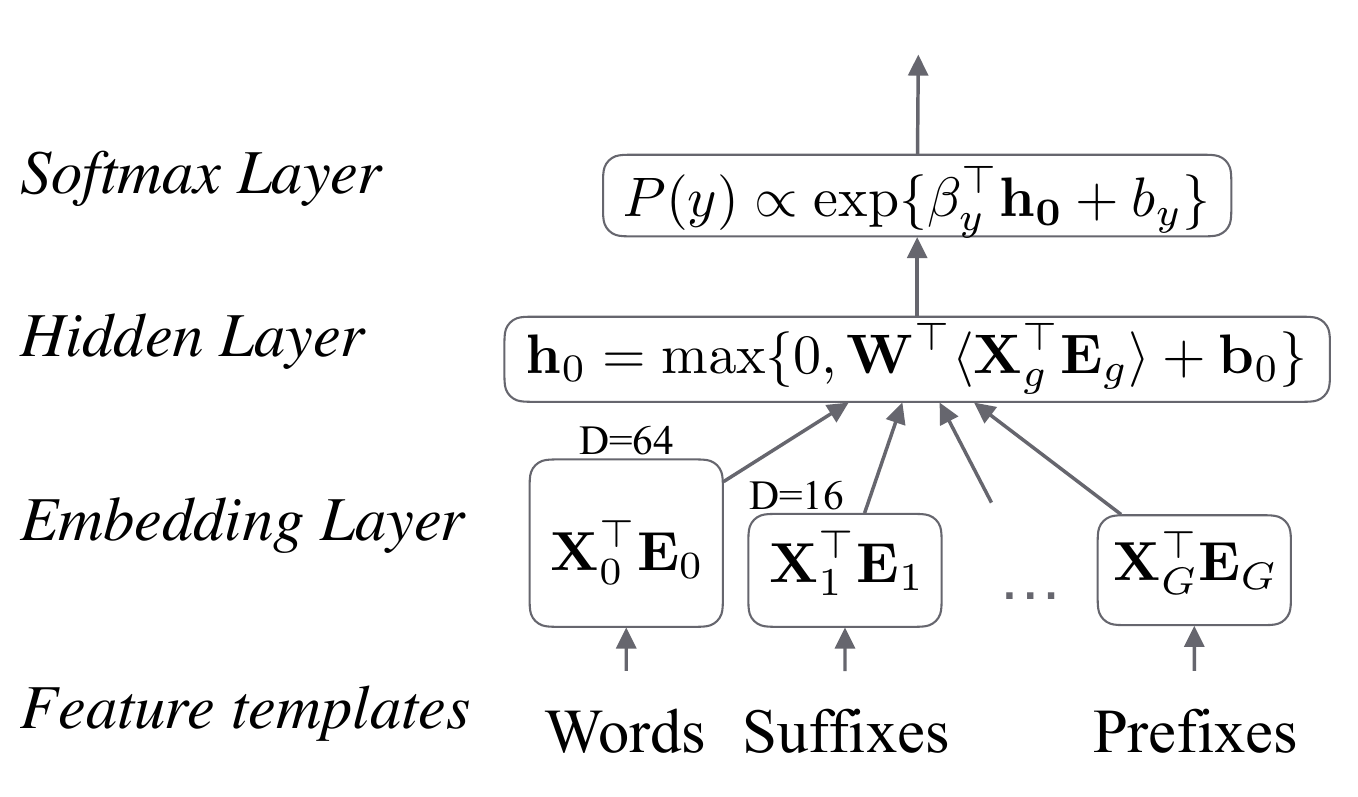}
  \caption{
    Elementary NN unit used in our model. Feature matrices from
    multiple channels are embedded, concatenated together, and fed into a
    rectified linear hidden layer. In the parser network, the feature inputs are
    continuous representations from the tagger network's hidden layer.}
  \label{fig:basic_unit}
\end{figure}

In this section, we introduce a novel neural network model for parsing and
tagging that incorporates POS tags as a regularization of learned implicit
representations.
The basic unit of our model (Figure \ref{fig:basic_unit}) is a simple,
feed-forward network that has been shown to work very well for parsing tasks
\cite{chen2014fast,weiss-etAl:2015:ACL}.
The inputs to this unit are feature matrices which are embedded and passed as
input to a hidden layer.
The final layer is a softmax prediction.

We use two such networks in this work: a window-based version for
tagging and a transition-based version for dependency parsing.
In a traditional stacking (pipeline) approach, we would use the
discrete predicted POS tags from the tagger as features in the parser
\cite{chen2014fast}.
In our model, we instead feed the continuous hidden layer activations of
the tagger network as input to the parser.
The primary strength of our approach is that the parser has access to all of the
features and information used by the POS tagger during training time, but it is
allowed to make its own decisions at test time.

To implement this, we show how we can re-use feature templates from
\newcite{chen2014fast} to specify the feed-forward connections from the tagger
network to the parser network.
An interesting consequence is that because this structure is a function of the
derivation produced by the parser, the final feed-forward structure of the
stacked model is not known until {\em run-time}.
However, because the connections for any specific parsing decision are fixed
given the derivation, we can still extract examples for training off-line by
unrolling the network structure from gold derivations. 
In other words, we can utilize our approach with the same simple stochastic
optimization techniques used in prior works.
Figure \ref{fig:unrolled} shows a fully unrolled architecture on a simple
example.

%
%
%


\subsection{The Tagger Network}

As described above, our POS tagger follows the basic structure from prior work
with embedding, hidden, and softmax layers.
Like the ``window-approach'' network of \newcite{collobertJMLR2011}, the tagger
is evaluated per-token, with features extracted from a window of tokens
surrounding the target.
The input consists of a rich set of features for POS tagging that are
deterministically extracted from the training data.
As in prior work, the features are divided into {\em groups} of different sizes
that share an embedding matrix $\mathbf{E}$.
Features for each group $g$ are represented as a sparse matrix $\mathbf{X}^g$
with dimension $F^g\times V^g$, where $F^g$ is the number of feature templates
in the group, and $V^g$ is the vocabulary size of the feature templates.
Each row of $\mathbf{X}^g$ is a one-hot vector indicating the appearance of each
feature.

The network first looks up the learned embedding vectors for each feature and
then concatenates them to form the embedding layer.
This embedding layer can be written as:
\begin{equation}
  \label{eq:concat-layer}
  \mathbf{h}_0 = [\mathbf{X}^g\mathbf{E}^g \mid \forall g]
\end{equation}
where $\mathbf{E}^g$ is a learned $V^g \times D^g$ embedding matrix for feature
group.
Thus, the final size $|\mathbf{h}_0| = \sum_g F^g D^g$ is the sum of all
embedded feature sizes.
The specific features and their dimensions used in the tagger are listed in
Table \ref{tab:tagger-features}.
Note that for all features, we create additional {\em null} value that triggers
when features are extracted outside the scope of the sentence.
We use a single hidden layer in our model and apply rectified linear unit (ReLU)
activation function over the hidden layer outputs.
A final softmax layer reads in the activations and outputs probabilities for
each possible POS tag.

 \begin{table}[t]
   \centering
   \begin{tabular}{ccc}
     Features ($g$) & Window & $D$ \\
     \hline
     Symbols & 1 & 8 \\
     Capitalization & +/- 1 & 4 \\
     Prefixes/Suffixes ($n=2,3$) & +/- 1 & 16 \\
     Words & +/-3 & 64
   \end{tabular}
   \caption{Window-based tagger feature spaces. ``Symbols'' indicates whether the word contains a hyphen, a digit or a punctuation.}
   \label{tab:tagger-features}
 \end{table}


\subsection{The Parser Network}

The parser component follows the same design as the POS tagger with the
exception of the features and the output space.
Instead of a window-based classifier, features are extracted from an
arc-standard parser configuration\footnote{Note that the ``stack'' in the parse
  configuration is separate from the ``stacking'' of the POS tagging network and
  the parser network (Figure \ref{fig:multitask}).} $c$ consisting of the stack
$s$, the buffer $b$ and the so far constructed dependencies \cite{Nivre:2004}.
Prior implementations of this model used up to four groups of discrete features: words,
labels (from previous decisions), POS tags, and morphological attributes
\cite{chen2014fast,weiss-etAl:2015:ACL,albert-etAl:2015:EMNLP}.

In this work, we apply the same design principle but we use an {\em implicitly}
learned intermediate representation in the parser to replace traditional discrete
features. We only retain discrete features over the labels in the
incrementally constructed tree (Figure \ref{fig:schema}).
Specifically, for any token of interest, we feed the hidden layer of the tagger
network evaluated for that token as input to the parser.
We implement this idea by re-using the feature templates from prior work as
indexing functions.

We define this process formally as follows.
Let $f_i(c)$ be a function mapping from parser configurations $c$ to indices in
the sentence, where $i$ denotes each of our feature templates.
For example, in Figure \ref{fig:schema}(a), when $i=${\tt stack$_0$}, $f_i(c)$
is the index of ``fox'' in the sentence.
Let $\bh_1^{tagger}(j)$ be the hidden layer activation of the tagger network
evaluated at token $j$.
We define the input $\bX^{implicit}$ by concatenating these tagger activations
according to our feature templates:
\begin{equation}
  \label{eq:concat-input}
  \bx_i^{implicit} \triangleq \bh^{tagger}_1(f_i(c)).
\end{equation}

Thus, the feature group $\bX^{implicit}$ is the row-concatenation of the hidden
layer activations of the tagger, as indexed by the feature templates.
We have that $F^{implicit}$ is the number of feature templates, and
$V^{implicit} = H^{tagger}$, the number of possible values is the number of
hidden units in the tagger.
Just as for other features, we learn an embedding matrix $\mathbf{E}^{implicit}$
of size $H^{implicit} \times F^{implicit}$.
Note that as in the POS tagger network, we reserve an additional {\em null}
value for out of scope feature templates.
A full example of this lookup process, and the resulting feed-forward network
connections created, is shown for a simple three-feature template consisting of
the top two tokens on the stack and the first on the buffer in Figure
\ref{fig:unrolled}.
See Table \ref{tab:tagger-features} for the full list of 20 tokens that we extract for each state.

\section{Learning with Stack-propagation}
\label{sec:learning}

In this section we describe how we train our stacking architecture.
At a high level, we simply apply backpropagation to our proposed continuous
form of stacking (hence ``stack-propagation.'')
There are two major issues to address: (1) how to handle the dynamic
many-to-many connections between the tagger network and the parser network, and
(2) how to incorporate the POS tag labels during training.

Addressing the first point turns out to be fairly easy in practice: we simply
unroll the gold trees into a derivation of (state, action) pairs that produce
the tree.
The key property of our parsing model is that the {\em connections} of the
feed-forward network are constructed incrementally as the parser state is
updated.
This is different than a generic recurrent model such as an LSTM, which passes
activation vectors from one step to the next.
The important implication at training time is that, unlike a recurrent network,
the parser decisions are conditionally independent given a fixed history.
In other words, if we unroll the network structure ahead of time given the gold
derivation, we do not need to perform inference when training with respect to
these examples.
Thus, the overall training procedure is similar to that introduced in
\newcite{chen2014fast}.

\begin{table*}[t]
  \centering
  \small
  \setlength\tabcolsep{2.2pt}
  \begin{tabular}{lcccccccccccccccccccc}
    \toprule
    Method & ar & bg & da & de & en & es & eu & fa & fi & fr & hi & id & it & iw & nl & no & pl & pt & sl & AVG \\
    \midrule
    \multicolumn{21}{l}{\textsc{No tags}} \\
    ~~B'15 LSTM &
    75.6 & {\bf 83.1} & 69.6 & {\bf 72.4} & 77.9 & 78.5 & 67.5 & 74.7 & {\bf 73.2} & 77.4 & 85.9 & {\bf 72.3} & {\bf 84.1} & 73.1 & 69.5 & 82.4 & {\bf 78.0} & {\bf 79.9} & {\bf 80.1} & 76.6 \\
    ~~Ours (window)  &
    {\bf 76.1} & 82.9 & {\bf 70.9} & 71.7 & {\bf 79.2} & {\bf 79.3} & {\bf 69.1} & {\bf 77.5} & 72.5 & {\bf 78.2} & {\bf 87.1} & 71.8 & 83.6 & {\bf 76.2} & {\bf 72.3} & {\bf 83.2} & 77.8 & 79.0 & 79.8 & {\bf 77.3} \\
    \midrule 
    \multicolumn{21}{l}{\textsc{Universal tagset}} \\
    ~~B'15 LSTM &
    74.6 & 82.4 & 68.1 & 73.0 & 77.9 & 77.8 & 66.0 & 75.0 & 73.6 & 78.0 & 86.8 & 72.2 & 84.2 & 74.5 & 68.4 & 83.3 & 74.5 & 80.4 & 78.1 & 76.2\\
    ~~Pipeline $P_{tag}$ &
    73.7 & 83.6 & 72.0 & 73.0 & 79.3 & 79.5 & 63.0 & 78.0 & 66.9 & 78.5 & 87.8 & 73.5 & 84.2 & 75.4 & 70.3 & 83.6 & 73.4 & 79.5 & 79.4 & 76.6 \\
    ~~RBGParser &
    75.8 & 83.6 & {\bf 73.9} & 73.5 & 79.9 & 79.6 & 68.0 & {\bf 78.5} & 65.4 & 78.9 & 87.7 & {\bf 74.2} & 84.7 & {\bf 77.6} & 72.4 & 83.9 & 75.4 & {\bf 81.3} & 80.7 & 77.6 \\          
    ~~Stackprop &
    {\bf 77.0} & {\bf 84.3} & 73.8 & {\bf 74.2} & {\bf 80.7} & {\bf 80.7} & {\bf 70.1} & {\bf 78.5} & {\bf 74.5} & {\bf 80.0} & {\bf 88.9} & 74.1 & {\bf 85.8} & 77.5 & {\bf 73.6} & {\bf 84.7} & {\bf 79.2} & 80.4 & {\bf 81.8} & {\bf 78.9} \\
    \bottomrule
  \end{tabular}
  \caption{
    Labeled Attachment Score (LAS) on Universal Dependencies Treebank.
    {\em Top:} Results without any POS tag observations.
    ``B'15 LSTM'' is the character-based LSTM model \cite{BallesterosDS15}, while ``Ours (window)'' is our window-based architecture variant without stackprop.
    {\em Bottom:} Comparison against state-of-the-art baselines utilizing the POS tags.
Paired t-tests show that the gain of Stackprop over all other approaches is significant ($p < 10^{-5}$ for all but RBGParser, which is $p < 0.02$).
  }
  \label{tab:ud-final}
\vspace{-0.5em}
\end{table*}


To incorporate the POS tags as a regularization during learning, we take a
fairly standard approach from multi-task learning.
The objective of learning is to find parameters $\Theta$ that maximize the data
log-likelihood with a regularization on $\Theta$ for both parsing and tagging:
\begin{multline}
  \label{eq:objective}
  \max_{\Theta} \lambda \sum_{\bx,y \in \mathcal{T}} \log(P_\Theta(y \mid \bx)) + \\
  \sum_{c,a \in \mathcal{P}} \log\left(P_\Theta(a \mid c)\right),
\end{multline}
where $\{\bx,y\}$ are POS tagging examples extracted from individual tokens and
$\{c,a\}$ are parser (configuration, action) pairs extracted from the unrolled
gold parse tree derivations, and $\lambda$ is a trade-off parameter.

We optimize this objective stochastically by alternating between two updates:
\begin{itemize*}
\item \textsc{Tagger:} Pick a POS tagging example and update the tagger network
  with backpropagation.
\item \textsc{Parser:} (Figure \ref{fig:schema}) Given a parser configuration
  $c$ from the set of gold contexts, compute both tagger and parser
  activations. Backpropagate the parsing loss through the stacked architecture
  to update both parser and tagger, ignoring the tagger's softmax layer
  parameters.
\end{itemize*}
While the learning procedure is inspired from multi-task learning---we only
update each step with regards one of the two likelihoods---there are subtle
differences that are important.
While a traditional multi-task learning approach would use the final layer of
the parser network to predict both POS tags and parse trees, we predict POS tags
from the first hidden layer of our model (the ``tagger'' network) only.
We treat the POS labels as regularization of our parser and simply discard the
softmax layer of the tagger network at test time.
As we will show in Section \ref{sec:experiments}, this regularization leads to
dramatic gains in parsing accuracy.
Note that in Section \ref{sec:discussion}, we also show experimentally that
stack-propagation is more powerful than the traditional multi-task approach, and
by combining them together, we can achieve better accuracy on both POS and
parsing tasks.


\subsection{Implementation details}

Following ~\newcite{weiss-etAl:2015:ACL}, we use mini-batched averaged
stochastic gradient descent (ASGD) \cite{bottou2010large} with momentum
\cite{hinton2012practical} to learn the parameters $\Theta$ of the network.
We use a separate learning rate, moving average, and velocity for the tagger
network and the parser; the \textsc{Parser} updates all averages, velocities,
and learning rates, while the \textsc{Tagger} updates only the tagging factors.
We tuned the hyperparameters of momentum rate $\mu$, the initial learning rate
$\eta_0$ and the learning rate decay step $\gamma$ using held-out data.
The training data for parsing and tagging can be extracted from either the same
corpus or different corpora; in our experiments they were always the same.

To trade-off the two objectives, we used a random sampling scheme to perform 10 epochs
of \textsc{Parser} updates and 5 epochs of \textsc{Tagger} updates.
In our experiments, we found that pre-training with \textsc{Tagger} updates for
one epoch before interleaving \textsc{Parser} updates yielded faster training
with better results.
We also experimented using the \textsc{Tagger} updates solely for initializing
the parser and found that interleaving updates was crucial to obtain
improvements over the baseline.




\section{Experiments}
\label{sec:experiments}

In this section, we evaluate our approach on several dependency
parsing tasks across a wide variety of languages.

\subsection{Experimental Setup}

We first investigated our model on 19 languages from the Universal Dependencies
Treebanks v1.2.\footnote{http://universaldependencies.org} We selected the 19
largest currently spoken languages for which the full data was freely available.
We used the coarse universal tagset in our experiments with no explicit
morphological annotations.
To measure parsing accuracy, we report unlabeled attachment score (UAS) and
labeled attachment score (LAS) computed on all tokens (including punctuation),
as is standard for non-English datasets.

For simplicity, we use the arc-standard \cite{Nivre:2004} transition system
with greedy decoding. 
Because this transition system only produces projective trees, we first apply a
projectivization step to all treebanks before unrolling the gold derivations
during training. We make an exception for Dutch, where we observed a
significant gain on development data by introducing the \textsc{Swap} action
\cite{nivre:2009:ACL} and allowing non-projective trees.

For models that required predicted POS tags, we trained a window-based tagger
using the same features as the tagger component of our stacking model.
We used 5-fold jackknifing to produce predicted tags on the training set.
We found that the window-based tagger was comparable to a state-of-the-art CRF
tagger for most languages.
For every network we trained, we used the development data to evaluate a small
range of hyperparameters, stopping training early when UAS no longer improved on
the held-out data.
We use $H=1024$ hidden units in the parser, and $H=128$ hidden units in the
tagger.
The parser embeds the tagger activations with $D=64$.
Note that following \newcite{BallesterosDS15}, we did not use any auxiliary data
beyond that in the treebanks, such as pre-trained word embeddings.

For a final set of experiments, we evaluated on the standard Wall Street Journal
(WSJ) part of the Penn Treebank \cite{marcus:1993:CL}), dependencies generated
from version 3.3.0 of the Stanford converter \cite{stanford_dependencies}.
We followed standard practice and used sections 2-21 for training, section 22
for development, and section 23 for testing.
Following \newcite{weiss-etAl:2015:ACL}, we used section 24 to tune any
hyperparameters of the model to avoid overfitting to the development set.
As is common practice, we use pretrained word embeddings from the
\texttt{word2vec} package when training on this dataset.


\begin{table}
  \centering
  \scalebox{1.0}{
  \begin{tabular}{lccc}
    \toprule
    Method & UAS & LAS \\
    \midrule
    \multicolumn{3}{l}{\textsc{No tags}} \\
    \quad Dyer et al. (2015) & 92.70 & 90.30 \\
    \quad Ours (window-based) & {\bf 92.85} & {\bf 90.77} \\
    \midrule
    \multicolumn{3}{l}{\textsc{Universal tagset}} \\
    \quad Pipeline $(P_{tag}$) &92.52 & 90.50 \\
    \quad Stackprop & {\bf 93.23} & {\bf 91.30} \\
    \midrule
    \multicolumn{3}{l}{\textsc{Fine tagset}} \\
    \quad Chen \& Manning (2014) & 91.80 & 89.60 \\
    \quad Dyer et al. (2015) & 93.10 & 90.90 \\
    \quad Pipeline ($P_{tag}$) & 93.10 & 91.16 \\
    \quad Stackprop & {\bf 93.43} & {\bf 91.41} \\
    \midrule 
    Weiss et al. (2015) & 93.99 & 92.05 \\
    Alberti et al. (2015) & 94.23 & 92.36 \\
    \bottomrule
  \end{tabular}}
\caption{
  WSJ Test set results for greedy and state-of-the-art methods. 
  For reference, we show the most accurate models from  \newcite{albert-etAl:2015:EMNLP} and \newcite{weiss-etAl:2015:ACL}, which use a deeper model and beam search for inference.}
  \label{tab:wsj}
\end{table}


\subsection{Results}


We present our main results on the Universal Treebanks in Table
\ref{tab:ud-final}.
We directly compare our approach to other baselines in two primary ways.
First, we compare the effectiveness of our learned continuous representations
with those of \newcite{albert-etAl:2015:EMNLP}, who use the predicted
distribution over POS tags concatenated with word embeddings as input to the
parser.
Because they also incorporate beam search into training, we re-implement a
greedy version of their method to allow for direct comparisons of token
representations.
We refer to this as the ``Pipeline ($P_{tag}$)'' baseline.
Second, we also compare our architecture trained without POS tags as
regularization, which we refer to as ``Ours (window-based)''. 
This model has the same architecture as our full model but with no POS
supervision and updates.
Since this model never observes POS tags in any way, we compare against a
recurrent character-based parser \cite{BallesterosDS15} which is
state-of-the-art when no POS tags are provided.\footnote{We thank
  \newcite{BallesterosDS15} for their assistance running their code on the
  treebanks.}
Finally, we compare to RGBParser \cite{lei2014low}, a state-of-the art
graph-based (non-greedy) approach.

Our greedy stackprop model outperforms all other methods, including the
graph-based RBGParser, by a significant margin on the test set (78.9\% vs
77.6\%).
This is despite the limitations of greedy parsing.
Stackprop also yields a 2.3\% absolute improvement in accuracy compared to using
POS tag confidences as features (Pipeline $P_{tag}$).
Finally, we also note that adding stackprop to our window-based model improves
accuracy in {\em every} language, while incorporating predicted POS tags into
the LSTM baseline leads to occasional drops in accuracy (most likely due to
cascaded errors.)

\begin{table}[t]
  \centering
  \begin{tabular}{lccc}
    \toprule
    Model Variant & UAS & LAS & POS\\
    \midrule
    \multicolumn{3}{l}{\em Arc-standard transition system} \\
    \quad Pipeline ($P_{tag}$) & 81.56 & 76.55 & 95.14 \\
    \quad Ours (window-based) & 82.08 & 77.08 & -  \\
    \quad Ours (Stackprop) & {\bf 83.38} & {\bf 78.78} & - \\
    \midrule
    \multicolumn{3}{l}{\em Joint parsing \& tagging transition system} \\
    \quad Pipeline ($P_{tag}$)& 81.61 & 76.57 & 95.30 \\
    \quad Ours (window-based) & 82.58 & 77.76 & 94.92 \\
    \quad Ours (Stackprop) & {\bf 83.21} & {\bf 78.64} & {\bf 95.43} \\
    \bottomrule
  \end{tabular}
  \caption{
    Averaged parsing and POS tagging results on the UD treebanks for joint 
    variants of stackprop. Given the window-based architecture, stackprop leads to higher parsing accuracies than joint modeling (83.38\% vs. 82.58\%). }
  \label{tab:pos-loss}
\end{table}

\begin{table*}
  \centering
  \small
  \begin{tabular}{lcccc}
    \toprule
    Token & married by a {\bf judge}. & Don't {\bf judge} a book by &and {\bf walked} away satisfied & when I {\bf walk} in the door \\
    \midrule
    Neighbors &mesmerizing as a {\em rat}. & doesn't {\em change} the company's & tried, and {\em tried} hard & upset when I {\em went} to \\
    &A {\em staple!} & won't {\em charge} your phone & and {\em incorporated} into & I {\em mean} besides me \\
    &day at a {\em bar}, then go & don't {\em waste} your money & and {\em belonged} to the & I {\em felt} as if I \\
    \midrule
    Pattern & a [noun] & 'nt [verb] & and [verb]ed & I [verb] \\
    \bottomrule
  \end{tabular}
  \caption{
    Four of examples of tokens in context, along with the three most
    similar tokens according to the tagger network's activations, and the simple pattern exhibited.
    Note that this model was trained with the Universal tagset which
    does not distinguish verb tense.}
  \label{tab:nn-interpret}
\end{table*}

\section{Discussion}
\label{sec:discussion}

\paragraph{Stackprop vs. other representations.}
One unexpected result was that, even {\em without} the POS tag labels at
training time, our stackprop architecture achieves better accuracy than either
the character-based LSTM or the pipelined baselines (Table \ref{tab:ud-final}).
This suggests that adding window-based representations--which aggregate over many
features of the word and surrounding context--is more effective than increasing
the expressiveness of individual word representations by using character-based
recurrent models.
In future work we will explore combining these two complementary approaches.

We hypothesized that stackprop might provide larger gains over the pipelined
model when the POS tags are very coarse.
We tested this latter hypothesis on the WSJ corpus by training our model using
the coarse universal tagsets instead of the fine tagset (Table \ref{tab:wsj}).
We found that stackprop achieves similar accuracy using coarse tagsets as the
fine tagset, while the pipelined baseline's performance drops dramatically. 
And while stackprop doesn't achieve the highest reported accuracies on the WSJ,
it does achieve competitive accuracies and outperforms prior state-of-the-art
for greedy methods \cite{dyer2015transition}.

\paragraph{Stackprop vs. joint modeling.}
An alternative to stackprop would be to train the {\em final} layer of our
architecture to predict both POS tags and dependency arcs.
To evaluate this, we trained our window-based architecture with the integrated
transition system of \newcite{bohnet2012transition}, which augments the
\textsc{SHIFT} transition to predict POS tags.
Note that if we also apply stackprop, the network learns from POS annotations
twice: once in the \textsc{Tagger} updates, and again the \textsc{Parser}
updates.
We therefore evaluated our window-based model both with and without stack-propagation,
and with and without the joint transition system.


We compare these variants along with our re-implementation of the pipelined
model of \newcite{albert-etAl:2015:EMNLP} in Table \ref{tab:pos-loss}.
We find that stackprop is always better, even when it leads to ``double
counting'' the POS annotations; in this case, the result is a model that is
significantly better at POS tagging while marginally worse at parsing than
stackprop alone.



\begin{figure}[t]
  \centering
  \begin{subfigure}[b]{0.45\textwidth}
    \centering
    \includegraphics[width=\textwidth]{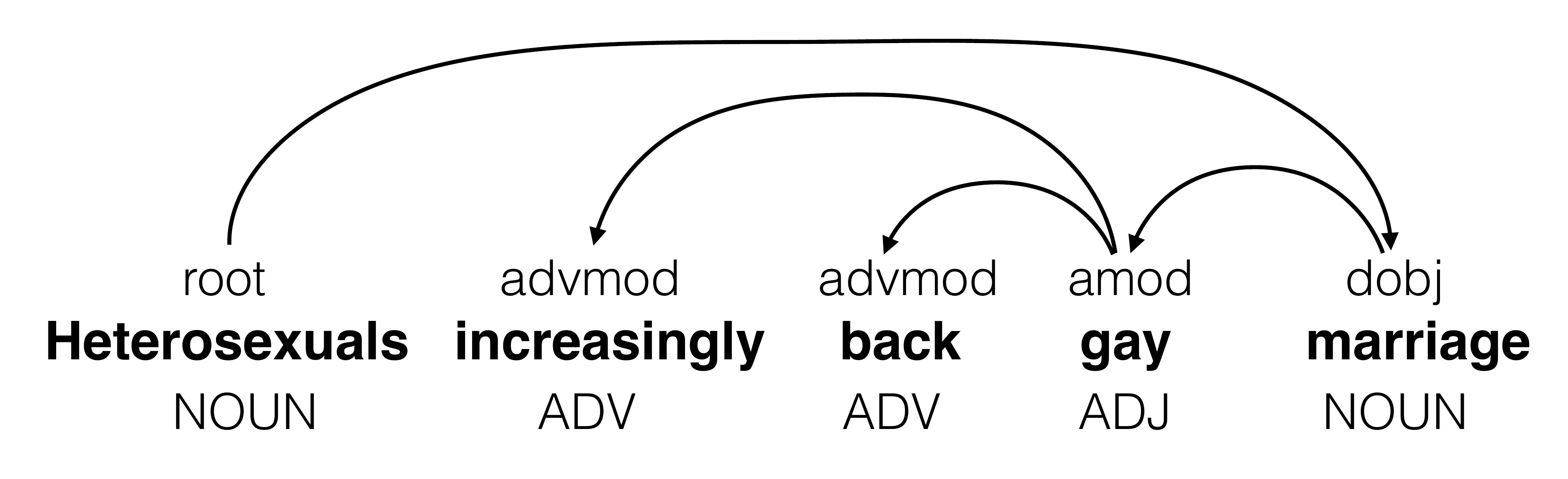}
    \caption{Tree by a pipeline model.}
  \end{subfigure}
  \begin{subfigure}[b]{0.45\textwidth}
    \centering
    \includegraphics[width=\textwidth]{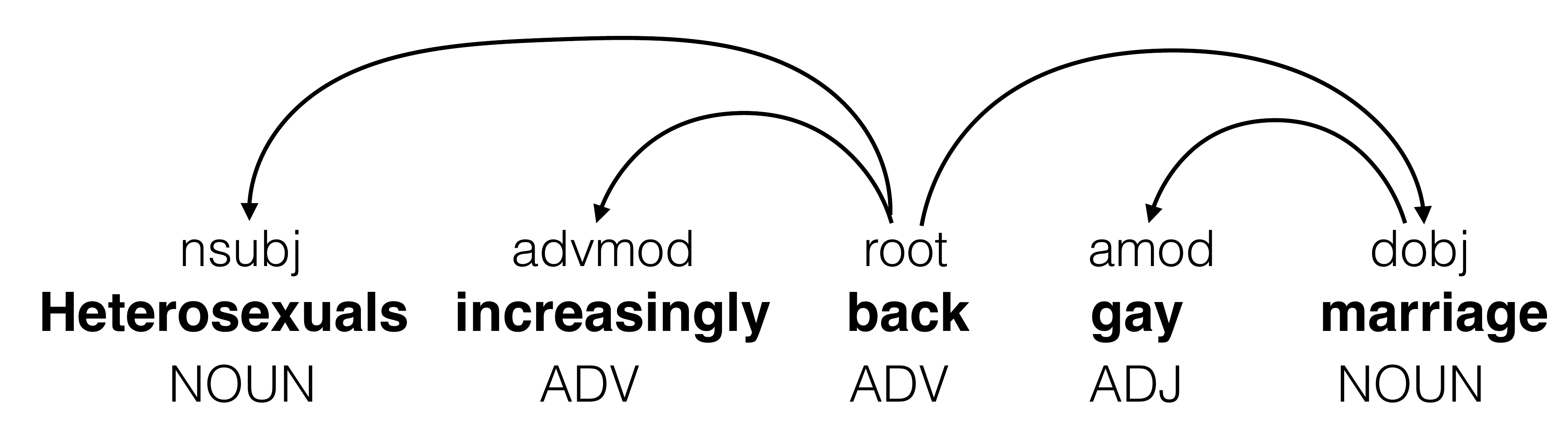}
    \caption{Tree by Stackprop model.}
  \end{subfigure}
  \caption{Example comparison between predictions by a pipeline model and a joint model. While both models predict a wrong POS tag for the word ``back'' (ADV rather than VERB), the joint model is robust to this POS error and predict the correct parse tree.
\vspace{-0.5em}
}\label{fig:case_study}
\end{figure}

\paragraph{Reducing cascaded errors.} As expected, we observe a significant
reduction in cascaded POS tagging errors. An example from the English UD
treebank is given in Figure \ref{fig:case_study}. Across the 19 languages in our
test set, we observed a 10.9\% gain (34.1\% vs. 45.0\%) in LAS on tokens where
the pipelined POS tagger makes a mistake, compared to a 1.8\% gain on the rest
of the corpora.

\paragraph{Decreased model size.} Previous neural parsers that use POS tags
require learning embeddings for words and other features on top of the
parameters used in the POS tagger \cite{chen2014fast,weiss-etAl:2015:ACL}.
In contrast, the number of total parameters for the combined parser and tagger
in the Stackprop model is reduced almost by half compared to the Pipeline model,
because the parser and tagger share parameters.
Furthermore, compared to our implementation of the pipeline model, we observed
that this more compact parser model was also roughly twice as fast.

\paragraph{Contextual embeddings.} Finally, we also explored the significance of
the representations learned by the tagger.
Unlike word embedding models, the representations used in our parser are
constructed for each token based on its surrounding context.
We demonstrate a few interesting trends we observed in Table
\ref{tab:nn-interpret}, where we show the nearest neighbors to sample tokens in
this contextual embedding space.
These representations tend to represent syntactic patterns rather than
individual words, distinguishing between the form (e.g. ``judge'' as a noun
vs. a verb') and context of tokens (e.g. preceded by a personal pronoun).

\section{Conclusions}

We present a stacking neural network model for dependency parsing and tagging.
Through a simple learning method we call ``stack-propagation,'' our model learns
effective intermediate representations for parsing by using POS tags as
regularization of implicit representations.
Our model outperforms all state-of-the-art parsers when evaluated on 19
languages of the Universal Dependencies treebank and outperforms other greedy
models on the Wall Street Journal.

We observe that the ideas presented in this work can also be as a principled
way to optimize up-stream NLP components for down-stream applications. 
In future work, we will extend this idea beyond sequence modeling to improve
models in NLP that utilize parse trees as features.
The basic tenet of stack-propagation is that the hidden layers of neural models
used to generate annotations can be used instead of the annotations themselves.
This suggests a new methodology to building deep neural models for NLP: we can
design them from the ground up to incorporate multiple sources of annotation and
learn far more effective intermediate representations.




\section*{Acknowledgments}
We would like to thank Ryan McDonald, Emily Pitler, Chris Alberti, Michael
Collins, and Slav Petrov for their repeated discussions, suggestions, and
feedback, as well all members of the Google NLP Parsing Team.
We would also like to thank Miguel Ballesteros for assistance running the
character-based LSTM.

\bibliography{paper}
\bibliographystyle{acl2016}

\end{document}